# Face Identification using Local Ternary Tree Pattern based Spatial Structural Components

Rinku Datta Rakshit[1], Dakshina Ranjan Kisku[2], Massimo Tistarelli[3] and Phalguni Gupta

[1] Department of Information Technology
Asansol Engineering College, Vivekananda Sarani,
Kanyapur, Asansol, West Bengal, 713305, India
[2] Department of Computer Science and Engineering
National Institute of Technology Durgapur, Mahatma Gandhi Road,
A-Zone ,West Bengal, 713209, India
[3] Computer Vision Lab, DAP, University of Sassari, Alghero, 07041 (SS), Italy
[4] Department of of Computer Science and Engineering,
IIT Kanpur, UP, 208016, India
{rakshit_rinku@rediffmail.com, drkisku@cse.nitdgp.ac.in, tista@uniss.it,
pg@cse.iitk.ac.in}

**Abstract.** This paper reports a face identification system which makes use of a novel local descriptor called Local Ternary Tree Pattern (LTTP). Exploiting and extracting distinctive local descriptor from a face image plays a crucial role in face identification task in the presence of a variety of face images including constrained, unconstrained and plastic surgery images. LTTP has been used to extract robust and useful spatial features which use to describe the various structural components on a face. To extract the features, a ternary tree is formed for each pixel with its eight neighbors in each block. LTTP pattern can be generated in four forms such as LTTP–Left Depth (LTTP-LD), LTTP–Left Breadth (LTTP-LB), LTTP–Right Depth (LTTP-RD) and LTTP–Right Breadth (LTTP-RB). The encoding schemes of these patterns are very simple and efficient in terms of computational as well as time complexity. The proposed face identification system is tested on six face databases, namely, the UMIST, the JAFFE, the extended Yale face B, the Plastic Surgery, the LFW and the UFI. The experimental evaluation demonstrates the most promising results considering a variety of faces captured under different environments. The proposed LTTP based system is also compared with some local descriptors under identical conditions.

Keywords: Face Identification, Local Descriptor, Ternary Tree, cosine similarity, sum of absolute differences, classifier.

## 1 Introduction

The outcomes of face recognition in the last decades exhibit enormous improvements in person identification while constrained face images are used with little variations.

However, it becomes difficult when a face identification system is presented with more twisted and unconstrained face images for identification. Further, plastic surgery images also make the identification more challenging. Now, face identification has moved on to unconstrained scenarios with more non-deterministic factors. Face recognition [7] is a continuing research process in computer vision and it has been attained significant attention due to extensive use in surveillance, law enforcement and information security. The wide applications of face recognition have been motivated the researchers due to its dynamicity in reliability and robustness. Recognizing a face in a controlled environment is not a difficult task, however, it raises an adverse situation in unconstrained scenarios. The success of face recognition relies on the choice of robust descriptors and features which can deal with uncertainties of a face image. Applying local descriptors to face recognition is a powerful approach and it has successfully used many face recognition systems. Among three modes of face recognition such as verification, identification and matching, identification seems to be a difficult mode where an unidentified probe face is compared with all face templates of the asserted identity in the database. In face identification, to find the identity of an unknown person, the input face image is compared with all face templates of the registered persons present in the database. The face identification is more demanding in crime inquisition, law enforcement, identification of a suspicious person in public places like school, bank, railway station, airport and border of a country. The main objective of this paper is to propose a robust face identification system which can deal with real life situations efficiently and at the same time it ensures the robustness of the system. To achieve this, a novel local descriptor LTTP is applied for extracting discriminatory facial features.

### 1.1 Related Works

To handle the challenging situations in face recognition, a large number of local descriptors [1, 2, 3, 5, 10, 12, 16, 18, 19] have been employed in many works. The existing local descriptors are mainly exploited for face verification, however, not experimented for face identification. Local descriptors have the competency to extract discriminatory and stable information subjected to extract the features information effectively from each structural component of a face image.

Some local descriptors such as Local Binary Pattern (LBP) [3, 12], Multi scale-Local Binary Pattern (MS-LBP) [10], Multi block-Local Binary Pattern (MB-LBP) [18], Local Texture Pattern (LTP) [16], Local Derivative Pattern (LDP) [19], Local Vector Pattern (LVP) [5], Local Graph Structure (LGS) [2] and Symmetric Local Graph Structure (SLGS) [1] are used for face recognition. Among them, LBP [3, 12] is known to be a widely used local descriptor to encode the local relationship of the pixels. The operator works with the eight neighbors of a pixel using the value of the center pixel as a threshold. A binary pattern is generated for every pixel of the face image and produces a transformed image (TI). Histogram generated from the transformed image (TI) is used as a feature vector. The LBP operator is able to extract micro-patterns from face images and these micro-patterns are invariant to monotonic grey scale transformations. The scale variation is an important factor which affects the face recognition system due to variations of texture in a different scale. To handle this situation, MS-LBP [10] is used to improve performance. The MS-LBP is able to extract micro-structures from the face images at different scales. Its accuracy is better than LBP operator due to enhance size of the neighborhood. The MB-LBP [18] is another variant which shows enhanced performance over LBP by computing the average value of a sub-region of the face image. LTP [16] uses a constant to threshold the pixels into three values. After thresholding, a ternary pattern is generated for every pixel and formed a transformed image. The LTP operator is found to be more discriminative and less sensory to noise in identical regions. On the contrary, LDP [19] uses the local derivative to encode directional pattern. To extract discriminatory features from a high-order derivative space, LVP can be used. To increase the robustness of the structure of micropatterns, LVP encodes different pair-wise directions of vectors and generate an invariant facial description. Some existing local descriptors which use graphs structures can also be found useful for providing compatible performance with LBP and its variants. Such as Local Graph Structures (LGS) [2] and Symmetric Local Graph Structure (SLGS) [1] are used local directional graph for a pixel to generate a binary pattern. These local descriptors have successfully been applied in face recognition due to their distinctive feature representation.

### 1.2 Major Contributions

The local descriptors are very powerful tools which can able to extract diverse and contrasting features from a face. The proposed work uses an efficient local descriptor called Local Ternary Tree Pattern (LTTP). It has the ability to identify the strength through the structural patterns which are derived from the face image. The descriptor is computationally not very expensive. Further, the identification performance under complicated environments is found to be outstanding. With structural uniqueness of LTTP, face identification can achieve its dominance phase.

The rest of the paper is organized as follows: Section 2 describes the proposed Local Ternary Tree Pattern (LTTP). Section 3 presents the proposed face identification framework which uses Local Ternary Tree Pattern in its feature extraction phase. The experimental results are provided in the next section. Finally, a conclusion is drawn in the last section.

## 2 Local Ternary Tree Pattern (LTTP)

The Local Ternary Tree Pattern (LTTP) is used effectively to extract the features which represent the invariant local textures of a face image in the way of extracting structural components on the face. Unlike LBP, the LTTP generates a ternary tree for each pixel of an image with its 8 neighbors and generates four patterns rather than one.

To extract the discriminatory and structural components from a face image, an image I(P) of size h×v is divided into a number of smaller sub-regions with dimension h̄×v̄, where h̄ <<h and v̄ <<v. Then LTTP operator is applied on each pixel of the gray scale face image and generates transformed value. To determine the transformed value for a pixel, a ternary tree is formed by considering the pixel as root of the tree covering eight neighbors in a 3×3 region. Then binary labeling of edges is performed. During labeling, the root node is compared with its child nodes and computes the differences between the root node and child nodes. If the difference is found positive or equal to 0, then assign 1 to the edge between the root node (source pixel) and child node (neighborhood pixel), otherwise assign 0 to that edge. Finally, binary labels (0 or 1) of edges of the ternary tree are concatenated together using four rules to form 8-bit binary pattern which is then converted to a decimal number and assigned to the target pixel. The four concatenation rules are: Left oriented–Depth First Traversal (LTTP-LD), Left oriented–Breadth First Traversal (LTTP-LB), Right oriented-Depth First Traversal (LTTP-RD), and Right oriented-Breadth First Traversal (LTTP-RB). A ternary tree with a target pixel '9' is shown in Figure 1. To encode the LTTP-LD pattern, at first starts from root node, goes to the depth in left side then goes to depth in right side and one by one concatenate all eight labels of edges. According to Figure 1, the sequence of edges to generate LTTP-LD is 9→9, 9→8, 9→6, 6→5, 6→11, 6→7, 9→8, 8→10 (A→B, B→E, A→C, C→F, C→G, C→H, A→D, D→I). To encode the LTTP-LB pattern, at first starts from root node, goes to breadth from left side to right side, covers all levels of the tree and one by one concatenate all eight labels of edges. According to Figure 1, the sequence of edges to generate LTTP-LB is 9→9, 9→6, 9→8, 9→8, 6→5, 6→11, 6→7, 8→10 (A→B, A→C, A→D, B→E, C→F, C→G, C→H, D→I). To encode the LTTP-RD pattern, at first starts from root node, goes to the depth in right side then goes to depth in left side and one by one concatenate all eight labels of edges. According to Figure 1, the sequence of edges to generate LTTP-RD is 9→8, 8→10, 9→6, 6→7, 6→11, 6→5, 9→9, 9→8 (A→D, D→I, A→C, C→H, C→G, C→F, A→B, B→E). Lastly, to encode the LTTP-RB pattern, at first starts from root node, goes to breadth from right side to left side, covers all levels of the tree and one by one concatenate all eight labels of edges. According to Figure 1, the sequence of edges to generate LTTP-RB is 9→8, 9→6, 9→9, 8→10, 6→7, 6→11, 6→5, 9→8 (A→D, A→C, A→B, D→I, C→H, C→G, C→F, B→E). The encoding scheme of LTTP considers the direct relationship of a target pixel with its neighbors as well as the relationship between the pixels that form the local ternary tree of the target pixel. This encoding scheme enables the LTTP to generate unique face representation and subsequently the improved identification performance.

To produce the decimal number by applying LTTP operator for a pixel $I(P_t)$, a binomial weight $2^q$ is multiplied to each label of edges of the ternary tree and then all multiplied labeled values are added.

The LTTP code for a target pixel $I(P_t)$ is given by

$$LTTP(I(P_t)) = \sum_{q=0}^{L-1} f(P_r - P_c) * 2^q \quad where \quad f(x) = \begin{cases} 1, & if \quad x \geq 0 \\ 0, & if \quad x < 0 \end{cases} \quad (1)$$

where $P_r$ denotes the gray value of root node (source pixel) and $P_c$ denotes the gray value of child node (neighboring pixel). In Equation (1), L is the total number of neighboring pixels of the target pixel. The pictorial representation of basic LTTP operator is shown in Figure 1.

The four different types of encoding schemes in LTTP would contribute a large number of distinguishable features to the feature set considering different environmental conditions including unconstrained and post-surgery.

## 3  Proposed Face Identification Framework

The aim of this paper is to present a robust face identification system which uses the proposed local descriptor – LTTP in its feature extraction phase to improve the overall performance of the system under degraded, unconstrained and post-surgery scenarios. The system consists of two main phases – feature extraction and classification.

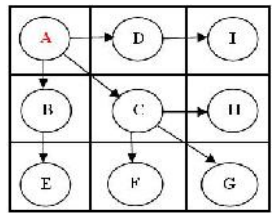
1(a)

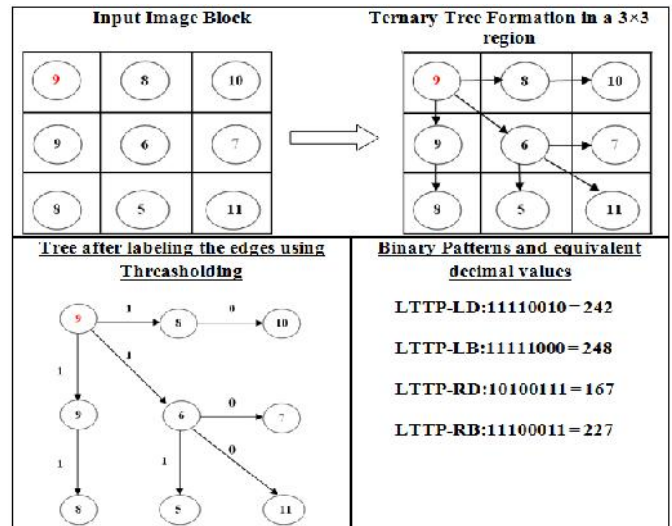
1(b)

**Fig. 1.** Pictorial Representation of Local Ternary Tree Pattern (LTTP).

The face identification [13, 14] is a paradigm of one to many matching that compares a probe image with all templates of face images present in a face database to infer the identity of the probe face. Suppose, a probe face image $I_p$ is considered for testing and a full set of gallery images G is considered for training. The feature vector $F(I_p)$ is generated for a probe image $I_p$. Then the objective function of the proposed face identification system can be defined as:

$$f(I_p) = \min_i(D(F(I_p), F(G_i))) \quad \text{where} \quad i = 1,2,3,\ldots\ldots,N \quad (2)$$

Where D(.) is the distance function used to measure the similarity between two feature vectors $F(I_p)$ and $F(G_i)$. $F(G_i)$ is the feature vector of $i^{th}$ gallery image.

The robust and discriminatory facial feature extraction is the key to the success of a face identification system. This phase transforms a raw face image into a transformed image and generates the feature vector. In this phase, the proposed Local Ternary Tree Pattern (LTTP) is applied on all pixels of a face image to generate the transformed image (TI) and finally, the feature vector is generated from the transformed image by flattening it. The proposed LTTP is capable to handle different challenges of a face image like changes in photometric condition, facial expression, head pose, facial accessories (makeup glasses etc.), imaging modality and unconstrained environment. An input face image and its transformed patterns generated using LTTP variants are shown in Figure 2.

The classification plays an important role in the proposed face identification system. Classification phase can be divided into two sub-phases such as matching and identity generation.

During matching, the template generated from a probe face image is compared with all templates produced from gallery face images and computes their similarity scores. The cosine similarity measure (CS) [11] and the sum of absolute differences (SAD) [4] are used to measure the similarity between the feature of the probe image and the feature of the gallery images.

During identity generation, the similarity scores produced by matching are used to generate a rank order list for each probe image. The gallery feature which has the maximum similarity to the probe feature is selected and the identity of the corresponding gallery image is considered as the identity of the probe image.

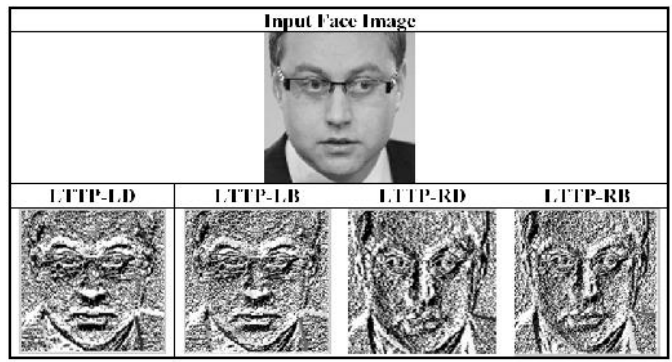

**Fig. 2.** An Input Image and Transformed Images generated by Applying LTTP Operator.

## 4  Evaluation

The proposed face identification system is evaluated on six challenging face databases, namely, the UMIST [8], the extended Yale face B [17], the JAFFE [16], the Plastic Surgery [15], the LFW [6] and the UFI [9] databases. Identification accuracy at Rank-1 strategy is determined using two distance metrics viz. cosine similarity measure (CS) [11] and the sum of absolute differences (SAD) [4]. During experiments, when the system takes a probe face as input, the Rank-1 strategy retrieves a face image corresponding to highest proximity, from database. In this experiment, the LTTP is used for facial feature extraction from the face image. However, prior to feature extraction, no image enhancement operation is applied. During training and testing, each face image is partitioned into a number of image blocks of size 3×3. Then, LTTP is applied to each block and obtain local textural and structural features which are concatenated together with other local features to form a global feature set. To generate probe and gallery sets, a random partition is used in the databases prior to training and testing. However, a different strategy other than random partition is adopted for the LFW and the UFI databases where a subset of images of the entire database is taken into consideration for evaluation and the subset is divided biasedly, to obtain the probe and gallery sets. Finally, the performance of the novel operator LTTP for face identification is compared with other local descriptors such as LBP [3, 12], LGS [2] and LTP [16] in identical scenarios with the same databases. A brief overview of six databases is listed in Table 1 and sample images of these databases are shown in Figure 3. The experimental results are shown in Table 2.

Table 1. Brief description of six face databases

| Name of Database | No. of Subjects | Total no. of Images | Image Format | Resolution | Description |
|---|---|---|---|---|---|
| JAFFE | 10 | 213 | .tiff | 121×146 | Facial Expression variations |
| UMIST | 20 | 564 | .pgm | 92×112 | Pose variations (profile to frontal) |
| Extended Yale face B | 38 | 2432 | .pgm | 40×46 | Illumination variations |
| LFW | 5749 | 13233 | .pgm | 64×64 | Unconstrained face images |
| UFI | 605 | 4921 | .pgm | 128×128 | Unconstrained face images |
| Plastic Surgery | 53 (obtained) 900 (original) | 106 (obtained) 1800 (original) | .jpg | 238×273 | Post-surgery variations |

The identification accuracy (IA) is computed using Equation (2) and Equation (3). For a probe image $I_p$, the Identification Accuracy 'IA' of the proposed system is defined as

$$IA = (\frac{1}{|M|} \sum_{p=1}^{|M|} \sum_{i=1}^{|N|} \Delta(\Phi(I_p), \Phi(G_i), R(I_p, G_i))) \times 100 \qquad (2)$$

where |M| is the size of probe set, |N| is the size of the gallery, and $\Delta(.)$ computes $k^{th}$ best match for the probe image $I_p$ as

$$\Delta(\Phi(I_p), \Phi(G_i), R(I_p, G_i)) = \begin{cases} 1, & if \quad \Phi(I_p) = \Phi(G_i) \quad and \quad R(I_p, G_i) = k \\ 0, & else \end{cases}$$
(3)

$\Phi(.)$ is a function which returns the class of an image and $R(I_p, G_i)$ returns the ranked position of a gallery image $G_i$ with respect to the test image $I_p$ from the probe set.

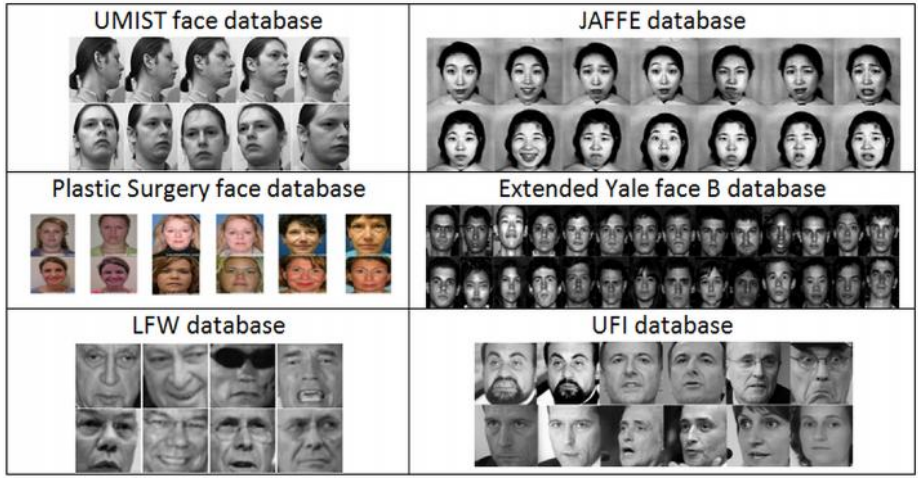

**Fig. 3.** Sample Images from the UMIST [8], the JAFFE [16], the extended Yale face B [17], the Plastic Surgery [15], the LFW [6] and the UFI [9] Face Databases.

### 4.1 Experimental Design and Protocol

Experiments are conducted with face images obtained from different databases and each of these databases is partitioned into two sets – probe and training. From the UMIST database [8] one image per subject (total 20 face images) having pose variation is considered in the probe set. Some of the subjects having eyeglasses are also considered in the probe set. Rest of the images are used for training. From the JAFFE database [16], one image per subject (total 10 face images) having varied facial expressions are included in the probe set and the remaining 203 face images are considered for the training set. In the extended Yale face B database [17], 8 images per subject (total 304 face images) with different illumination conditions are included in the probe set and remaining 2128 images are included in the training set. In the Plastic Surgery face database [15], 15 to 22 face images labeled as post-surgery are included in the probe set and 53 face images labeled as pre-surgery are considered for the training set. It is due to the difficulty in separation of pre-surgery and post-surgery images and unavailability of plastic surgery face images, only upto 53 images are considered for the experiment. The face images present in the LFW database [6] are captured in unconstrained environments. The accuracy is highly affected due to illumination variation, facial expression changes, occlusion (external or self), pose variations and facial accessories like scarf, mask, hat or sunglasses. The database contains 13233 face images of 5749 individuals. We conducted the experiment only on 1,680 subjects with the criteria of having two or more than two face images. One image per subject is considered in the probe set and the rest of the images are included in the training set. The UFI database [9] is another unconstrained database which contains face images of 605 subjects. The presence of various backgrounds in face images makes the task of identification more difficult. During the experiment on the database, several probe sets depending upon different variations present in face images are created and 4316 images are used in the training set.

### 4.2 Experimental Results

The UMIST [8] face database covers a range of poses from profile to frontal views. From experimental outcomes, it has been observed that the proposed LTTP based face identification gives 100% accuracy at Rank-1 using cosine similarity (CS) and the sum of absolute differences (SAD) while handling pose variations of face images. The experimental results reveal that the LTTP is robust to frontal as well as different pose variations of the face images.

The Extended Yale Face B database [17] containing face images having a number of illumination effects and it makes the database challenging. The identification accuracy determined on this database achieves 100% at Rank-1 using cosine similarity (CS) and the sum of absolute differences (SAD) while LTTP operator is used as a feature descriptor in face identification. The operator is capable to handle the effects of illumination variations of face images.

The identification accuracy determined on the JAFFE [16] database achieves 100% at Rank-1 using cosine similarity (CS) and the sum of absolute differences (SAD) while the database covers 7 facial expressions. The outcomes reveal that LTTP operator could capture variations in facial appearance due to expression changes very efficiently.

The post-surgery variations on a face create a challenging situation in identification. The Plastic Surgery face database [15] contains pre-surgery and post-surgery face images of 900 subjects. However, the evaluation uses images of 53 subjects as the rest of the images from 847 subjects are not found due to some difficulty. The proposed face identification system has achieved 100% accuracy at Rank-1 using cosine similarity (CS) and the sum of absolute differences (SAD) as classifiers.

The proposed face identification achieves 100% and 99.66% accuracies on the LFW [6] and the UFI [9] unconstrained face databases respectively while Rank-1 strategy is employed. The outcomes reveal that the proposed LTTP shows the robustness when face images are captured under unconstrained environments. The experimental results are shown in Table 2.

Table 2. Identification Accuracies (%) determined at Rank-1 on six face databases.

| Name of Databases | LTTP-LD | | LTTP-LB | | LTTP-RD | | LTTP-RB | |
|---|---|---|---|---|---|---|---|---|
| | CS | SAD | CS | SAD | CS | SAD | CS | SAD |
| UMIST | 100 | 100 | 100 | 100 | 100 | 100 | 100 | 100 |
| JAFFE | 100 | 100 | 100 | 100 | 100 | 100 | 100 | 100 |
| Extended YALE face B | 100 | 100 | 100 | 100 | 100 | 100 | 100 | 100 |
| Plastic Surgery | 100 | 100 | 100 | 100 | 100 | 100 | 100 | 100 |
| LFW | 100 | 100 | 100 | 100 | 100 | 100 | 100 | 100 |
| UFI | 99.66 | 99.66 | 99.66 | 99.66 | 99.66 | 99.66 | 99.66 | 99.66 |

Most of the works on the LFW database have been reported in face verification. Since face identification is a demanding task in law enforcement and in an unconstrained environment, a low-quality probe face may not be able to provide a sufficient description of a face image, therefore this motivates to check the effect of different environmental conditions such as pose, facial expression, lighting conditions, occlusions, background, accessories and photographic quality explicitly during experiments. The effect of different environmental conditions are analyzed separately in the LFW database. Several experiments are conducted on the LFW database to check the robustness of the proposed descriptor under different environmental conditions separately as the unconstrained environment is the collection of various situations that affect the appearance of a face image to a large extent. It is enlisted best identification accuracy at Rank-1 in Table 2 while a subset of images is selected biasedly from a number of subsets. The experimental outcomes reveal that the proposed LTTP operator shows the robustness in the probe set of the LFW database where mainly pose variations, facial expression variations and illumination variations are considered.

In law enforcement, to determine the identity of a subject based on one or more probe images, a top 200 ranked list may be retrieved from the gallery and which suffices for analyzers [20]. This motivates to conduct further experiments on the LFW database to test the combined effect of different environmental conditions. The experiments are conducted on the dataset containing twenty and more than twenty images per subject and generate identification accuracy up to Rank-50 shown in Table 3.

Table 3. Identification accuracies (%) on the LFW dataset having the criteria of twenty images per subject at Rank-1, Rank-10, Rank-20, Rank-30, Rank-40 and Rank-50.

| Rank | LTTP-LD | | LTTP-LB | | LTTP-RD | | LTTP-RB | |
|---|---|---|---|---|---|---|---|---|
| | CS | SAD | CS | SAD | CS | SAD | CS | SAD |
| 1 | 50.90 | 56.36 | 45.45 | 49.09 | 32.72 | 32.72 | 25.45 | 36.36 |
| 10 | 89.09 | 89.09 | 81.81 | 87.27 | 72.72 | 78.18 | 76.36 | 78.18 |
| 20 | 94.54 | 96.36 | 92.72 | 94.54 | 80 | 90.90 | 83.63 | 92.72 |
| 30 | 98.18 | 98.18 | 96.36 | 98.18 | 89.09 | 98.18 | 89.09 | 96.36 |
| 40 | 98.18 | 100 | 96.36 | 100 | 92.72 | 98.18 | 92.72 | 96.36 |
| 50 | 100 | 100 | 96.36 | 100 | 96.36 | 98.18 | 94.54 | 98.18 |

A number of experiments are conducted on the UFI face database and enlist the best identification accuracy at Rank-1 achieved among different probe sets of the database, in Table 2. During experiments, the effects of different environmental conditions like pose, facial expression, illumination, occlusion, image quality and background are experimented explicitly and analyze the effect of different conditions. If an image with pose variation is present in the probe set, but not an image with pose variations in the training set, then it is very difficult to identify the probe image at Rank-1. Presence of auxiliary background in the probe image or images captured in extreme illumination or self-occlusion or low-quality image affects the performance of the identification system adversely. The UFI is an unconstrained database which contains images captured with no restrictions over environmental conditions. To analyze the effect of different environmental conditions, different probe sets out of 605 subjects are created. The probe set which contains images with facial expression variations; illumination variations are identified correctly at Rank-1. The probe set which contains images with pose variations and self-occlusions are identified correctly if such kinds of variations are present in the training set. In these cases, 100% accuracy is achieved. If low-quality images and images captured in extreme illumination are present in the probe set, it affects the performance of the identification system adversely. In these cases, special preprocessing approaches are required before feature extraction.

To check the combined effect of different environmental conditions (unconstrained environment), further experiments are conducted on the UFI database in an unbiased way, where one image per subject (total 605 subjects) is considered together in a probe set and 4316 images in the training set. The experimental results are shown in Table 4.

Table 4. Identification accuracies (%) on the UFI dataset having images of all subjects in probe set at Rank-1, Rank-10, Rank-20, Rank-30, Rank-40, Rank-50.

| Rank | LTTP-LD | | LTTP-LB | | LTTP-RD | | LTTP-RB | |
|---|---|---|---|---|---|---|---|---|
| | CS | SAD | CS | SAD | CS | SAD | CS | SAD |
| 1 | 44.13 | 50.24 | 42.64 | 49.09 | 39.83 | 42.97 | 38.51 | 43.30 |
| 10 | 62.31 | 67.6 | 60.49 | 67.27 | 54.04 | 58.84 | 53.88 | 61.15 |
| 20 | 67.76 | 74.04 | 66.11 | 72.39 | 59.50 | 66.77 | 59.17 | 67.10 |
| 30 | 69.75 | 76.85 | 69.25 | 75.86 | 64.46 | 69.91 | 63.47 | 71.40 |
| 40 | 72.72 | 79.50 | 71.40 | 78.67 | 67.10 | 72.56 | 67.27 | 73.71 |
| 50 | 73.71 | 80.99 | 73.55 | 80.16 | 69.25 | 74.21 | 69.75 | 75.20 |

4.8 Comparative Study

This section presents a comparison of LTTP based system with some existing local descriptors such local binary pattern (LBP) [3, 12], local graph structure (LGS) [2] and local texture pattern (LTP) [16] which are simulated and tested on the same databases. The identification accuracy achieved by LTTP is compared with the accuracies obtained by LBP, LGS, and LTP at the corresponding ranks. Table 5 shows the comparison summary on six face databases.

When evaluation is performed on the JAFFE database, it can be seen from Table 5 that the proposed LTTP yields 100% accuracy at Rank-1 using both CS and SAD metrics, LBP produces 90% accuracy using the CS classifier and 100% accuracy using the SAD classifier, also, LGS achieves 100% accuracy using both the classifiers and LTP achieves 70% accuracy using both the classifiers. The experimental outcomes reveal that the proposed LTTP performs well in facial expression variations. When evaluation is performed on the UMIST face database, both LTTP and LBP operators achieve 100% accuracy using both the classifiers, the LGS achieves 100% accuracy using the SAD classifier and 95% accuracy using the CS classifier and LTP achieves 90% accuracy using the CS and 80% accuracy using the SAD classifier. Thus, the proposed descriptor performs well in case of pose variations. When the same identification system is experimented on the extended Yale face B database, the LBP achieves 62.5% accuracy using the CS and 74.67% accuracy using the SAD, the LGS achieves 70% accuracy using the CS and 76.97% accuracy using the SAD, the LTP achieves 38.15% accuracy using the CS and 44.07% accuracy using the SAD, while LTTP achieves 100% accuracy using both classifiers. The outcomes on the extended Yale face B database reveal that the proposed LTTP is more robust than existing local descriptors in case of illumination variations. When the same identification framework is evaluated on the Plastic Surgery face database, the proposed LTTP performs well. The LBP achieves 46.66% accuracy using the CS and 53.33% accuracy using the SAD, the LGS achieves 73.33% accuracy using both the CS and SAD classifiers, and the LTP achieves 13.33% accuracy using both the CS and SAD classifiers, while LTTP achieves 100% accuracy using both classifiers. When the same identification framework tested on the LFW face database the LBP achieves 17.77% accuracy using the CS and 15.55% accuracy using the SAD, the LGS achieves 28.88% accuracy using the CS and 31.11% accuracy using the SAD, the LTP achieves 4.44% accuracy using the CS and 6.66% accuracy using SAD, while LTTP achieves 100% accuracy using both CS and the SAD. When the UFI database is used in the experiment, the LBP achieves 47% accuracy using CS and 48% accuracy using SAD, the LGS achieves 50.5% accuracy using CS and 55.5% accuracy using SAD, the LTP achieves 10% using CS and 11.5% using SAD, while LTTP achieves 99.66% accuracy using both the CS and SAD classifiers. From Table 5 we can say that the proposed LTTP is much more robust than existing local descriptors in case of the pose variations, facial expression variations, illumination variations, post-surgery variations and unconstrained environments.

Table 5. Comparison of the proposed LTTP with state-of-the-art local descriptors at Rank-1 strategy.

| Name of face databases | Name of local descriptors | | | | | | | |
|---|---|---|---|---|---|---|---|---|
| | LBP | | LGS | | LTP | | LTTP | |
| | CS | SAD | CS | SAD | CS | SAD | CS | SAD |
| JAFFE | 90 | 100 | 100 | 100 | 70 | 70 | 100 | 100 |
| UMIST | 100 | 100 | 95 | 100 | 90 | 80 | 100 | 100 |
| Extended YALE face B | 62.5 | 74.67 | 70 | 76.97 | 38.15 | 44.07 | 100 | 100 |
| Plastic Surgery | 46.66 | 53.33 | 73.33 | 73.33 | 13.33 | 13.33 | 100 | 100 |
| LFW | 17.77 | 15.55 | 28.88 | 31.11 | 4.44 | 6.66 | 100 | 100 |
| UFI | | 47 | 48 | 50.5 | 55.5 | 10 | 11.5 | 99.66 99.66 |

## 5  Conclusion and Future Works

This paper has presented a promising work on face identification which has used a novel local descriptor (LTTP) for facial feature extraction. The proposed descriptor LTTP, as well as the face identification system, have validated through a variety of face images captured under extensive experiments including plastic surgery faces. The LTTP is a ternary tree based local descriptor where a pixel is represented by a ternary tree with its eight neighbors. Then using a thresholding rule a binary pattern is generated for the target pixel. The experiments illustrated that the use of LTTP in feature extraction phase of the face identification system has enabled it with greater discriminative power. The proposed LTTP is more robust than other local descriptors because its ternary tree structure is able to capture more discriminatory information from a face. The LTTP descriptor might have a huge impact if it is used for sustainable facial recognition systems. However, a few more tests on vibrant databases such as low resolution as well as heterogeneous face images will establish its usefulness as an unconquerable feature representation tool.

## References

1. Abdullah, M. F. A., Sayeed, M. S., Muthu, K. S., Bashier, H. K., Azman, A., & Ibrahim, S. Z.: Face recognition with symmetric local graph structure (slgs). *Expert Systems with Applications*, *41*(14), 6131-6137 (2014).
2. Abusham, E., & Bashir, H.: Face recognition using local graph structure (LGS). *Human-Computer Interaction. Interaction Techniques and Environments*, 169-175 (2011).
3. Ahonen, T., Hadid, A., & Pietikäinen, M.: Face recognition with local binary patterns. *Computer vision-eccv 2004*, 469-481(2004).
4. Alsaade, F.: Fast and accurate template matching algorithm based on image pyramid and sum of absolute difference similarity measure. *Research Journal of Information Technology*, *4*(4), 204-211(2012).
5. Fan, K. C., & Hung, T. Y.: A novel local pattern descriptor—local vector pattern in high-order derivative space for face recognition. *IEEE transactions on image processing*, *23*(7), 2877-2891(2014).
6. Huang, G. B., Mattar, M., Berg, T., & Learned-Miller, E.: Labeled faces in the wild: A database for studying face recognition in unconstrained environments. In: *Workshop on faces in'Real-Life'Images: detection, alignment, and recognition*. (2008).
7. Jain, A. K., & Li, S. Z.: *Handbook of face recognition*. New York: Springer (2011).
8. Kisku, D. R., Mehrotra, H., Gupta, P., & Sing, J. K.: Robust multi-camera view face recognition. *International Journal of Computers and Applications*, *33*(3), 211-219 (2011).
9. Lenc, L., & Král, P.: Unconstrained facial images: Database for face recognition under real-world conditions. In: *Mexican international conference on artificial intelligence 2015,* pp. 349-361. Springer, Cham (2015).

10. Chan, C. H., Kittler, J., & Messer, K.: Multi-scale local binary pattern histograms for face recognition. In: *International conference on biometrics 2007,* pp. 809-818. Springer, Berlin, Heidelberg (2007).
11. Nguyen, H. V., & Bai, L.: Cosine similarity metric learning for face verification. In: *Asian conference on computer vision* 2010, pp. 709-720. Springer, Berlin, Heidelberg (2010).
12. Ojala, T., Pietikainen, M., & Maenpaa, T.: Multiresolution gray-scale and rotation invariant texture classification with local binary patterns. *IEEE Transactions on pattern analysis and machine intelligence*, *24*(7), 971-987 (2002).
13. Rakshit, R. D., Nath, S. C., & Kisku, D. R.: An improved local pattern descriptor for biometrics face encoding: a LC–LBP approach toward face identification. *Journal of the Chinese institute of engineers*, *40*(1), 82-92 (2017).
14. Rakshit, R. D., Nath, S. C., & Kisku, D. R.: Face identification using some novel local descriptors under the influence of facial complexities. *Expert Systems with Applications*, *92*, 82-94 (2018).
15. Singh, R., Vatsa, M., Bhatt, H. S., Bharadwaj, S., Noore, A., & Nooreyezdan, S. S.: Plastic surgery: A new dimension to face recognition. *IEEE Transactions on Information Forensics and Security*, *5*(3), 441-448 (2010).
16. Suruliandi, A., Meena, K., & Rose, R. R.: Local binary pattern and its derivatives for face recognition. *IET computer vision*, *6*(5), 480-488 (2012).
17. UCSDRepository,2001http://vision.ucsd.edu/ leekc/ExtYaleDatabase/ExtYaleB.html
18. Zhang, L., Chu, R., Xiang, S., Liao, S., & Li, S. Z.: Face detection based on multi-block lbp representation. In: *International conference on biometrics* 2007, pp. 11-18. Springer, Berlin, Heidelberg (2007).
19. Zhang, B., Gao, Y., Zhao, S., & Liu, J.: Local derivative pattern versus local binary pattern: face recognition with high-order local pattern descriptor. *IEEE transactions on image processing*, *19*(2), 533-544 (2010).
20. Best-Rowden, L., Han, H., Otto, C., Klare, B. F., & Jain, A. K. (2014). Unconstrained face recognition: Identifying a person of interest from a media collection. *IEEE Transactions on Information Forensics and Security*, *9*(12), 2144-2157.